\documentclass[runningheads]{llncs}

\usepackage{eccv}

\usepackage{amsmath}
\usepackage{multirow}
\usepackage{booktabs}
\usepackage[table]{xcolor}
\usepackage{geometry}
\definecolor{lightgreen}{RGB}{222,239,218}
\definecolor{lightblue}{RGB}{210,230,245}
\definecolor{rowgray}{gray}{0.85}
\usepackage{graphicx}
\usepackage{algorithm}
\usepackage{algpseudocode}
\usepackage{wrapfig}
\usepackage{booktabs}
\usepackage{wrapfig}

\usepackage{eccvabbrv}

\usepackage{graphicx}
\usepackage{booktabs}

\usepackage[accsupp]{axessibility}  
\usepackage{marvosym}

\usepackage{hyperref}

\usepackage{orcidlink}

\usepackage{caption}

\begin{document}

\title{ELDiff: When Evidential Learning Meets Text-to-Image Diffusion} 

\titlerunning{ELDiff}

\author{Qingtao Pan\inst{1,2}\orcidlink{0009-0009-8939-9559} \and
Kai Ye\inst{2} \and
Zhihao Dou\inst{2}\orcidlink{0000-0002-3525-7442} \and
Bing Ji\inst{1}\textsuperscript{\href{mailto:b.ji@sdu.edu.cn}{\raisebox{-0.15ex}{\scriptsize(\Letter)}}}\,\orcidlink{0000-0003-1326-4120} \and
Shuo Li\inst{2}\orcidlink{0000-0002-5184-3230}} 

\authorrunning{Q.~Pan et al.}

\institute{Shandong University, Jinan, Shandong, China \\
\email{b.ji@sdu.edu.cn}\\ \and
Case Western Reserve University, Cleveland, OH, USA}

\maketitle

\begin{abstract}
In multi-object text-to-image (T2I) diffusion, ensuring semantic consistency between textual prompts and generated visual content is crucial for image synthesis. However, such consistency constraint is often underemphasized in the denoising process of diffusion models. Although token supervised diffusion models can mitigate this issue by learning object-wise consistency between the image content and object segmentation maps, it tends to suffer from the problems of segmentation map bias and semantic overlap conflict, especially when involving multiple objects. In this paper, we propose ELDiff, a new evidential learning-supervised T2I diffusion model, which leverages the advantages of uncertainty metric and conflict detection to enhance the fault tolerance of unreliable segmentation maps and suppress semantic conflicts, strengthening object-wise consistency learning. Specifically, a pixel evidence loss is proposed to restrain overconfidence in unreliable labels through evidential regularization, and a token conflict loss is designed to weaken the contradiction between semantics through optimizing a measured conflict factor. Extensive experiments show that our ELDiff outperforms existing training based and train-free based T2I diffusion models on SD v1.4, SD v2.1, SDXL, SD v3.5, and Qwen-Image, without requiring additional inference-time manipulations. Notably, ELDiff can be seamlessly extended to the existing training pipeline of T2I diffusion models. Code can be found at \url{https://github.com/QingtaoPan/ELDiff}.
\end{abstract}

\section{Introduction}
\label{sec:intro}

Recent advances in text-to-image (T2I) diffusion models \cite{b1,b2,b3,b6,b7} have demonstrated remarkable progress within computer vision applications, propelling the capabilities of general generative models to a new level. Leveraging their exceptional generation ability, these models enable users to synthesize highly realistic and creative visual content through user-specified text prompts. However, they often struggle to compose multiple objects into a coherent scene since the noise is predicted by a full text prompt in the denoising objective of T2I diffusion models \cite{b8,b9}, which requires the model to be equipped with the ability of understanding both the full text prompt and individual linguistic concepts from the prompt.

Several studies \cite{b8,b9,b10,b11,b12,b13} have addressed the challenge of generating images including multiple objects. They enable the generated images better reflect the text prompts through utilizing fine-grained text information. As proposed in TokenCompose \cite{b8}, the fundamental idea is to perform consistency constraint between the text prompts and the image contents in the finetuning stage by imposing training objectives at the token level. It adopts Grounding DINO \cite{b14} and Segment Anything (SAM) \cite{b15} to obtain the grounding segmentation map of each text token for local supervision of each text token. Once fine-tuned, the model can generate images by composing different combinations of words from text prompts in the inference stage. Although these methods have shown great success, they still suffer from segmentation map bias and semantic overlap conflict, which is problematic particularly when the generated image involves multiple objects.

We hypothesize that the limitations come from two aspects. On the one hand, the masks (the segmentation maps) generated directly by SAM are not always reliable \cite{b16}, which results in overfitting to spurious regions. On the other hand, semantic overlaps result in conflicts \cite{b17} among different text tokens since the consistency constraint between each noun token from the text prompt and its corresponding segmentation map is individually optimized. Therefore, we seek to address the following problems: \textit{How to cope with segmentation map bias and semantic overlap conflict for more effective consistency constraint in T2I diffusion models?}

\textit{Motivation:} Inspired by the above observation, we find it desirable to tackle such problems by considering the uncertainty metric of pixel-level predictions and quantifying the semantic overlap conflict. Evidential Deep Learning (EDL) \cite{b18,b19}, which is able to collect evidence of each category and estimate the epistemic uncertainty in a single forward pass. Given this, we consider each pixel of each noun token from the text prompt of as a classification unit and provide evidence values per class to reflect both the confidence and the associated uncertainty. In this way, the model tends to produce low-confidence predictions in mislabeled regions of the biased segmentation map, thereby avoiding overfitting to incorrect supervision. Moreover, Dempster-Shafer Evidence Theory (DST) \cite{b20}, an evidence combination rule, allows beliefs from different evidences to be combined to obtain a new belief \cite{b21,b22}. It is able to quantify the conflict level through introducing the conflict penalty item in overlapping areas. Inspired by this, we consider each noun token from the text prompt as different evidences and utilize DST combination rule to measure their conflict. Therefore, the model can perceive overlapping conflicts between different semantic evidences, effectively alleviating the semantic competition problem caused by independent optimization.

To this end, we propose ELDiff, an end-to-end fine-tuning strategy to enhance multi-object composition through a novel consistency constraint (i.e., EDL driven object-wise consistency constraint) between the text prompts and the image contents. Specifically, a pixel evidence loss is proposed to cope with the segmentation map bias. It models the per-token cross-attention maps as evidence distributions (i.e., Dirichlet distribution) and calculates the cross entropy using the mean value of the predicted Dirichlet distribution instead of directly using point predictions. In this way, when the uncertainty is high (i.e., the distribution is loose), the loss of EDL will not excessively penalize the model, allowing the model to remain cautious when facing biased segmentation map and avoid being forced to learn incorrectly. In addition, a token conflict loss is proposed to alleviate semantic overlap conflict. It regards per-token cross-attention map as multiple evidence embeddings and utilizes DST combination rule to aggregate such multiple evidence embeddings for conflict measurement. The measured conflict is utilized as an optimization target, enabling the model to adaptively adjust the spatial distribution of different tokens' cross-attention maps, thus mitigating semantic conflicts arising from attention map overlap. Our main contributions are summarized as the following:

\begin{itemize}
    \item We formulate‌ the problems of segmentation bias and semantic overlap conflict during the object-wise consistency constraint between the text prompts and the image contents. And we propose a new T2I finetuning pipeline, ELDiff, which introduces EDL based pixel-level uncertainty metric and DST based token-level conflict quantification to address these two problems. ELDiff can be seamlessly integrated into the existing T2I diffusion models.
    
    \item We design a pixel evidence loss to restrain overconfidence in unreliable labels through simultaneously considering pixel-level classification prediction and uncertainty estimation.

    \item We propose a token conflict loss to weaken the semantic overlap conflict through treating the conflict factor measured by the DST combination rule as the optimization objective.

    \item Extensive comparisons with previous outstanding methods demonstrate that ELDiff effectively improves the performance in composing multiple objects without additional inference cost.
\end{itemize}

\section{Related Work}
\textbf{Text-to-Image Synthesis.} The field of text-to-image synthesis \cite{b23,b24,b25,b26,b27,b28} has demonstrated remarkable generative power in various fields, such as text-to-image generation \cite{b29,b30}, text-to-video generation \cite{b31,b32}, and text-to-3D generation \cite{b33,b34}. By encoding text prompts into a condition vector using a pretrained CLIP \cite{b35}, T2I diffusion models have achieved successful applications in image generation. Although progress has been made, compositional generation still struggles when text prompts involve multiple objects and intricate relationships \cite{b36}. One potential method to generate multi-target images is manipulating the latent and cross-attention maps \cite{b10,b37,b12,b9,b38,b8,b40,b13}. More recently, TokenCompose \cite{b8} optimizes the cross-attention maps based on segmentation maps to provide dense consistency constraint. It achieves strong compositionality and image quality without additional inference time. Despite its superiority, it suffers from the segmentation bias and optimizing multiple objects individually leads to semantic overlap conflicts, resulting in degraded image generation quality. In this work, we present a different approach to address these problems for more reliable multi-category image composition through consistency constraint with evidential supervision.

\noindent
\textbf{Evidential Deep Learning.} EDL is gradually developed and refined based on Dempster-Shafer theory of evidence \cite{b20} and Subjective Logic theory \cite{b41}. The core idea of EDL is to collect evidence of each category and construct a Dirichlet distribution parameterized over the collected evidence to model the distribution of class probabilities. In addition to the probability of each category, the predictive uncertainty can be quantified from the distribution by Subjective Logic theory in the forward pass. EDL has been applied in a variety of research areas, e.g., EDL based classification \cite{b42,b43} and segmentation \cite{evivlm,textbcs}, evidential models for regression \cite{b44,b45}, adversarial robustness \cite{b46} and calibration \cite{b47}. Most existing EDL approaches typically incorporate evidential loss along with evidence regularization to guide the uncertainty behavior \cite{b48}, \cite{b49} of the evidence. In this work, we focus on EDL loss for segmentation and DST for multi-evidence aggregation. It is a promising way to tackle the problem of segmentation map bias and semantic overlap conflict, enabling more effective image synthesis involving multiple objects.

\section{Method}
\begin{figure*}[t]
\centering
\includegraphics[width=0.8\linewidth]{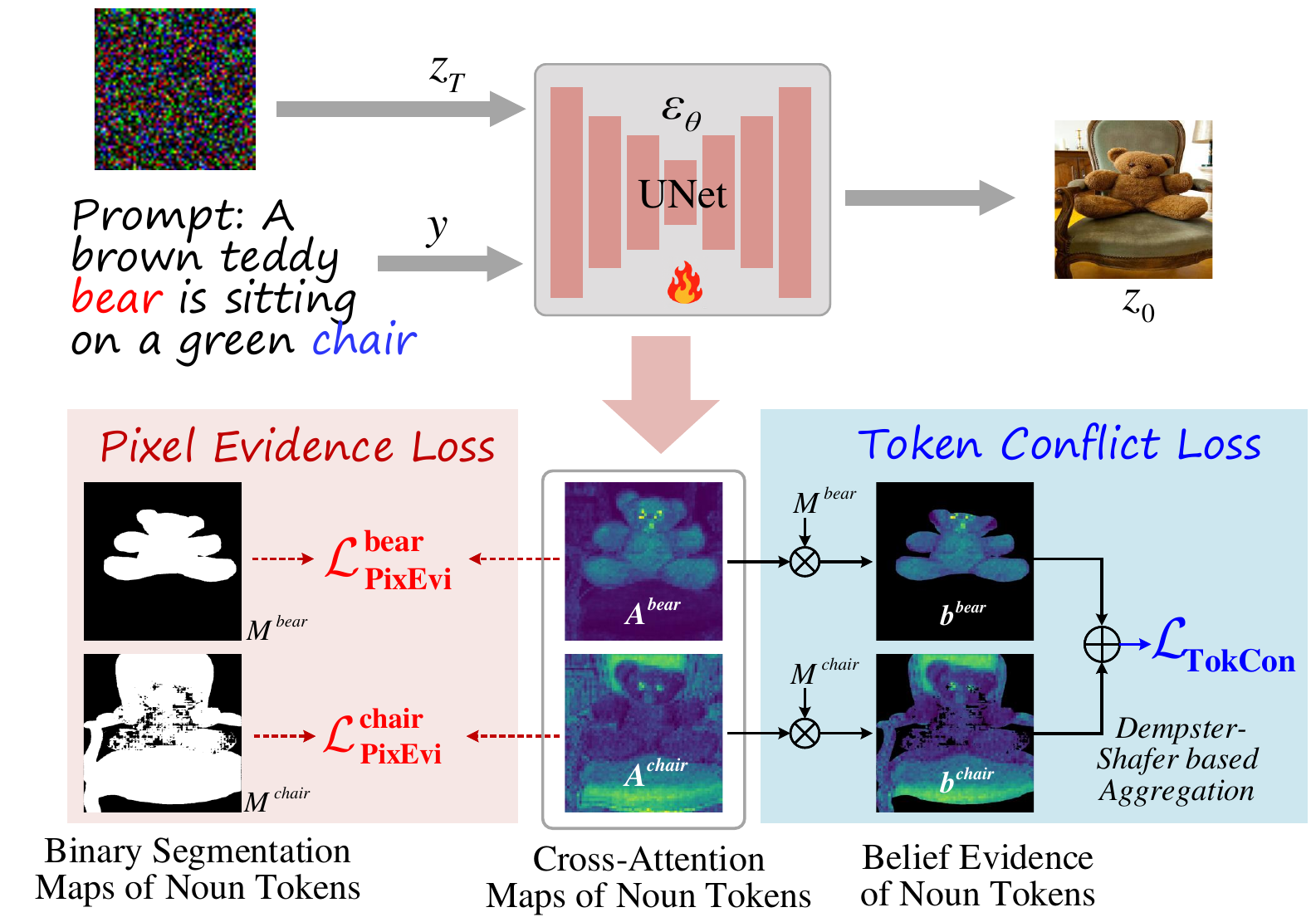}
\caption{\textbf{Overview of ELDiff.} The main components include: (1) The pixel evidence loss drives reliable consistency constraint between the text prompts and the image contents through pixel-level uncertainty estimation; (2) The token conflict loss weakens the contradiction between semantics.}
\label{fig2}
\end{figure*}

Our ELDiff (Fig. \ref{fig2}) proposes a pixel evidence loss to restrain overconfidence in unreliable labels when conducting pixel-level supervision between the image content and the object segmentation map, and designs a token conflict loss to weaken the contradiction between semantics through optimizing a measured conflict factor. In Section \ref{sec3.1}, we illustrate the problem setup of the segmentation map bias and semantic overlap conflict. Subsequently, in Section \ref{sec3.2}, we detail how the pixel evidence loss address this through pixel-level uncertainty metric. Finally, in Section \ref{sec3.3}, we demonstrate how the token conflict loss alleviates the semantic conflict through DST combination rule.

\begin{figure*}[t]
\centering
\includegraphics[width=\linewidth]{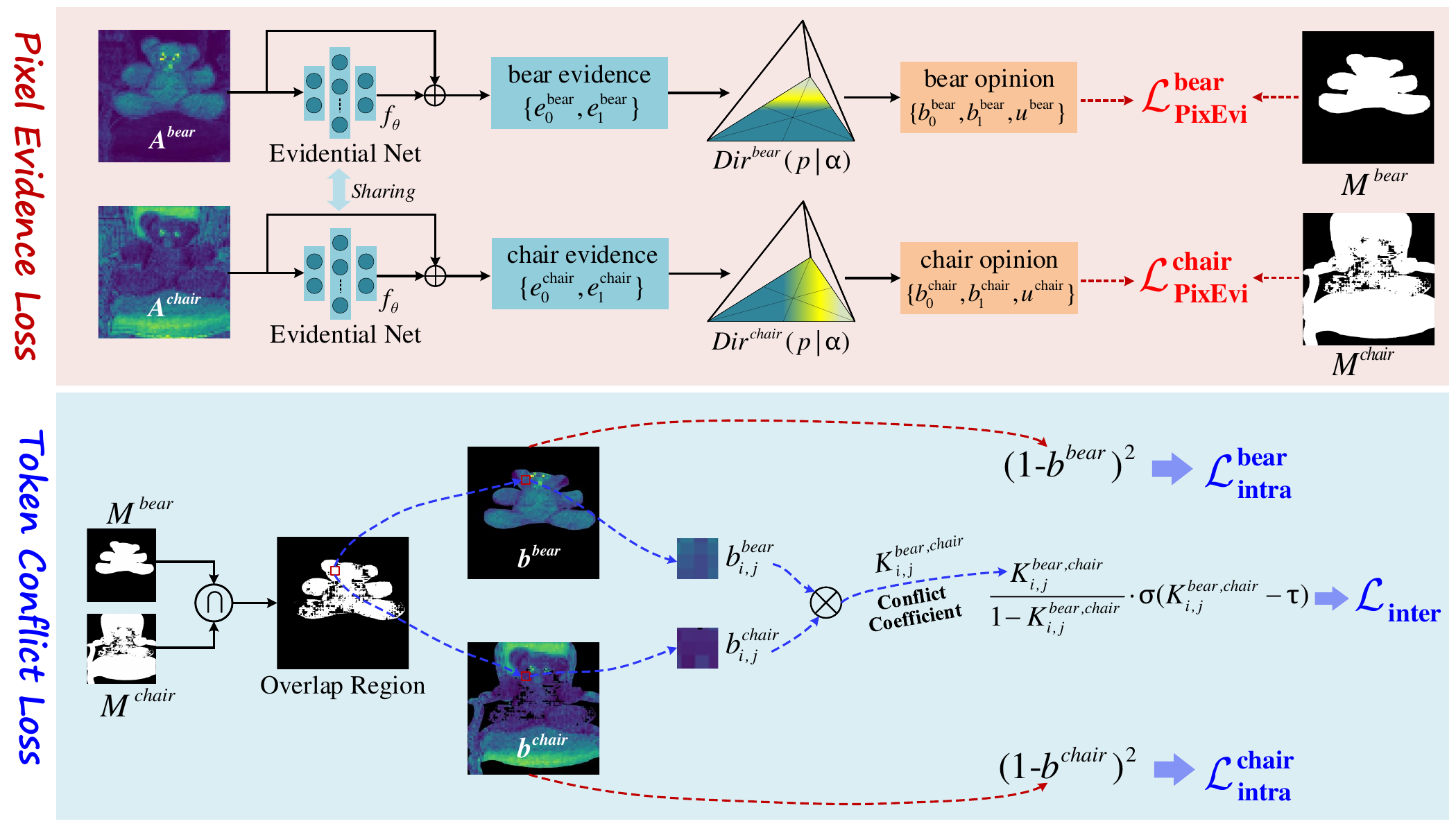}
\caption{\textbf{Detailed diagram of pixel evidence loss and token conflict loss.} For pixel evidence loss, the noun token's cross-attention maps are fed into the evidential Net to generate noun token's evidences. Then, these evidences are used to build the Dirichlet distribution, outputting pixel-level probabilities and the uncertainty for reliable pixel-level supervision. For token conflict loss, the conflict coefficient of the overlapping area between noun token's cross-attention maps is calculated for optimization. Moreover, the activations of different noun tokens are aggregated into respective corresponding spatial regions.}
\label{fig3}
\end{figure*}

\subsection{Problem setup}
\label{sec3.1}
\textbf{Segmentation map bias.} While consistency constraint based T2I diffusion methods have shown great success \cite{b8,b52}, the generation effect heavily depends on the quality of the segmentation map used for consistency constraint. For example, when performing a training constraint between the text token guided cross-attention map and the corresponding binary segmentation map, the model overfits to incorrect regions if binary segmentation map is biased. We mathematically formulate this problem in \color{red}Appendix B.1\color{black}.

\noindent
\textbf{Semantic overlap conflict.} Although $\mathcal{L}_{pixel}$ can focus the cross-attention map of nouns in text prompts on the target area, a side effect of this loss is that separately optimizing each cross-attention map is contradictory when different semantics overlap \cite{b53}. To be specific, this per-noun supervision strategy inherently assumes that the visual concepts of different nouns are spatially disjoint. For example, in the phrase “a cat on a chair,” the noun “cat” and “chair” may correspond to overlapping regions in the image due to physical interaction or spatial proximity. Supervising each noun’s cross-attention map with an independent segmentation map leads to semantic overlap conflicts, where the same pixel may be simultaneously assigned to multiple concepts. This introduces contradicting gradients during training, impairing the model’s ability of concept-to-region mappings. We mathematically formulate this problem in \color{red}Appendix B.2\color{black}.

\subsection{Pixel Evidence Loss}
\label{sec3.2}

To address the segmentation bias, our key insight is to add uncertainty metric between the cross-attention map and the segmentation map to achieve more reliable consistency constraint. We accomplish this by leveraging the epistemic uncertainty ability of EDL, which can quantify the credibility of its predictions. Instead of treating the attention score of each pixel as a deterministic prediction, we reinterpret it as evidence supporting its foreground or background class. As shown in Fig. \ref{fig3}, these evidence values are modeled via a Dirichlet distribution to construct an uncertainty-aware attention learning process.

\textit{Evidence-based Attention Modeling.} Given a cross-attention map $A$ and the corresponding binary segmentation map $M$, $A$ is fed into an MLP evidence network $f_{\phi}(\cdot)$ with an activation function (Softplus) to map $A$ into the evidence space, thus outputting evidence values $e$. For each pixel $A_{i,j}$, its evidence is defined as $e_{i,j}\in\mathbb{R}_{\geq0}$, from which the parameters of a Dirichlet distribution $D(p\mid\alpha)$ are obtained as $\alpha_{i,j}=e_{i,j}+1$. The Dirichlet distribution $D(p\mid\alpha)$ is considered as the conjugate prior to the multinomial distribution. It provides a predictive distribution for the segmentation results, defned as follows:
\begin{equation}
\left.D(p\mid\alpha)=\left\{\begin{array}{ll}\frac{1}{\beta(\alpha)}\prod_{k=1}^Kp_k^{\alpha_k-1}&for \quad p\in\Omega\\0&otherwise\end{array}\right.\right.
\label{eq1}
\end{equation}
where $p=[p_1,\ldots p_k]$ are the parameters of the multinomial distribution, $\beta(\alpha)$ is the high-dimensional multinomial beta function, and $\Omega$ is the $K-1$ dimensional unit simplex, defned as:
\begin{equation}\Omega=\left\{p\left|\sum_{c=1}^Cp^c=1\mathrm{~and~}0\leq p^1,\ldots,p^C\leq1\right.\right\}
\label{eq2}
\end{equation}
Subsequently, Subjective Logic \cite{b40} is applied for the optimization of Dirichlet distribution parameter optimization. It establishes a theoretical foundation linking Dirichlet distribution parameters to confidence and uncertainty quantification. Specifcally, for the predicted cross-attention map $A$, it provides a mass belief and uncertainty, satisfying:
\begin{equation}
u_{i,j}^c+\sum_{c=1}^Cb_{i,j}^c=1
\label{eq3}
\end{equation}
where $b_{i,j}$ and $u_{i,j}^c$ are the cross-attention map probability of
the pixel $(i,j)$ for the $c$-th class and the uncertainty of the pixel $(i,j)$, respectively. The the mass of belief and uncertainty for the pixel $(i,j)$ can be expressed as follows:
\begin{equation}
b_{i,j}^{c}=\frac{e_{i,j}^{c}}{S}=\frac{\alpha_{i,j}^{c}-1}{S}\quad\mathrm{and}\quad u_{i,j}=\frac{C}{S}
\label{eq4}
\end{equation}
where $S=\sum_{c=1}^C\left(e_{i,j}^c+1\right)$ is the Dirichlet strength. It describes that the higher the allocated belief mass, the more evidence is obtained for pixel $(i,j)$. Conversely, the less evidence obtained, the greater the overall uncertainty for the segmentation of pixel $(i,j)$.

Considering that, on the simplex, the ideal Dirichlet distribution should concentrate its mass on the vertex corresponding to the true class label. The distribution parameters should be close to 1 for all incorrect classes and significantly larger for the correct one. Therefore, it is necessary to design a loss function that guides the model to optimize the Dirichlet parameters in a way that minimizes segmentation uncertainty. Specifcally, We use the cross-entropy loss $\mathcal{L}_{ce}=\sum_{c=1}^{C}-y^{c}\log{(p^{c})}$ to link the Dirichlet distribution with the belief representation in subjective logic. Based on the evidence, this loss can be reformulated to reflect the expected prediction under the Dirichlet distribution, enabling the model to learn both accurate and uncertainty-aware segmentations, defned as follows:
\begin{equation}
\begin{split}
\mathcal{L}_{ice} &= 
\int \left[\sum_{c=1}^{C}-y^{c}\log(p^{c})\right]
\frac{1}{\beta(\alpha)}
\prod_{c=1}^{C}\left(p^{c}\right)^{\alpha^{c}-1}dp \\
&= 
\sum_{c=1}^C y^c \left(
\psi(S) - \psi(\alpha^c)
\right)
\end{split}
\label{eq5}
\end{equation}
where $y^c$ is the ground truth labels and $\psi\left(\cdot\right)$ is digamma function. We further incorporate a Kullback-Leibler (KL) divergence loss to suppress evidence for incorrect classes while preventing the Dirichlet parameter of the ground-truth class from being forced to 1, defined as:
\begin{equation}
\begin{split}
\mathcal{L}_{KL} &=
\log\left(
\frac{
\Gamma\left(\sum_{c=1}^C \tilde{\alpha}^c\right)
}{
\Gamma(C)\sum_{c=1}^C \Gamma\left(\tilde{\alpha}^c\right)
}
\right) \\
&\quad
+ \sum_{c=1}^{C} \left(\tilde{\alpha}^{c} - 1\right)
\left[
\psi\left(\tilde{\alpha}^{c}\right)
- \psi\left(\sum_{c=1}^{C}\tilde{\alpha}^{c}\right)
\right]
\end{split}
\label{eq6}
\end{equation}

where $\tilde{\boldsymbol{\alpha}}^c=y^c+(1-y^c)\odot\boldsymbol{\alpha}^c$ and $\Gamma\left(\cdot\right)$ is the gamma function.

The loss $\mathcal{L}_\mathrm{{PixEvi}}$ in this section consists of cross-entropy loss $\mathcal{L}_{ce}$, $\mathcal{L}_{ice}$, and $\mathcal{L}_{KL}$.
\begin{equation}
\mathcal{L}_\mathrm{{PixEvi}}=\frac{1}{N}\sum_{n=1}^{N}(\lambda_1\mathcal{L}_{ce}^{(n)}+\lambda_2(\mathcal{L}_{ice}^{(n)}+\mathcal{L}_{KL}^{(n)}))
\label{eq7}
\end{equation}
where $n$ represents each text token that belongs to a noun within the text prompt, and $\lambda_1$, $\lambda_2$, and $\lambda_3$ are hyperparameters to balance these three losses. The cross-entropy loss $\mathcal{L}_{ce}$ is adopted to maximize the consistency between the cross-attention map and the binary segmentation map.

\subsection{Token Conflict Loss}
\label{sec3.3}
To solve the semantic overlap conflict, we propose to aggregate all noun-level cross-attention maps using Dempster-Shafer Evidence Theory (DST) \cite{b20} to explicitly quantify inter-noun conflicts (Fig. \ref{fig3}), guiding the model to learn semantically decoupled attention regions. 

\textit{Cross-Attention Map Aggregation.} Given a set of per-noun cross-attention maps $\{A^1,A^2,\ldots,A^N\}$, we construct basic belief assignment function based on DST.
\begin{equation}
b^{n}=\gamma^nA^n \cdot M^n
\label{eq9}
\end{equation}
where $b^{n}$ is the belief evidence belonging to the $n$-th noun category. $\gamma^n\in(0,1)$ represents learnable factor. $M^n$ is the corresponding binary segmentation map. For tow nouns $(v,w)$, the conflict coefficient within the overlapping area $M^{v}\cap M^{w}$ is calculated by:
\begin{equation}
K^{v,w}_{i,j} =
\begin{cases}
b^{v}_{i,j} \cdot b^{w}_{i,j}, & \text{if} (i,j) \in M^v \cap M^w \\
0, & \text{otherwise}
\end{cases}
\label{eq10}
\end{equation}
Subsequently, we define a intra-object consistency loss $\mathcal{L}_{\mathrm{intra}}$ to aggregate its activations of the cross-attention map into certain subregions of its target regions and a inter-object conflict loss $\mathcal{L}_{\mathrm{inter}}$ to punish conflicting activations in overlapping areas, defined as:
\begin{equation}
\mathcal{L}_\mathrm{intra}=\sum_n(1-b^n)^2
\label{eq11}
\end{equation}
\begin{equation}
\mathcal{L}_{\mathrm{inter}}=\sum_{(i,j)\in M^v\cap M^w}\frac{K^{v,w}_{i,j}}{1-K^{v,w}_{i,j}}\cdot\sigma(K^{v,w}_{i,j}-\tau)
\label{eq12}
\end{equation}
where $\sigma$ is Sigmoid function and $\tau$ is constraint threshold. The final token conflict loss is the sum of $\mathcal{L}_\mathrm{intra}$ and $\mathcal{L}_{\mathrm{inter}}$, defines as:
\begin{equation}
\mathcal{L}_{\mathrm{TokCon}}=\frac{1}{N}\sum_{n=1}^{N}(\eta_1\mathcal{L}_\mathrm{intra}^{(n)}+\eta_2\mathcal{L}_{\mathrm{inter}}^{(n)})
\label{eq13}
\end{equation}
Finally, our ELDiff is jointly optimized by $\mathcal{L}_{LDM}$, $\mathcal{L}_{\mathrm{ELDiff}}$, and $\mathcal{L}_{\mathrm{TokCon}}$. The training objective is $\mathcal{L}_{\mathrm{ELDiff}}=\mathcal{L}_{LDM}+\mathcal{L}_{\mathrm{PixEvi}}+\mathcal{L}_{\mathrm{TokCon}}$

\begin{table*}[t]
\caption{Evaluation results of our method against other methods based on SD v1.4. ELDiff holistically demonstrates the best performance regarding Multi-category Instance Composition, Photorealism, Realism and Efficiency. The best score is in blue , with the second-best score in green. (C) denotes the COCO instance validation set and (F) denotes the Flickr30K instance validation set.}
\centering
\small
\setlength{\tabcolsep}{5pt}
\renewcommand{\arraystretch}{1.1}
\resizebox{\textwidth}{!}{
\begin{tabular}{lcccccccccc}
\toprule
\multicolumn{1}{c}{\multirow{2}[2]{*}{\textbf{Model}}} & \multicolumn{5}{c}{\textbf{Multi-category Instance Composition}} & \multicolumn{4}{c}{\textbf{Photorealism}} & \textbf{Efficiency} \\
\cmidrule(lr){2-6} \cmidrule(lr){7-10} \cmidrule(lr){11-11}
& \textbf{OA}↑ & \textbf{MG2}↑   & \textbf{MG3}↑  & \textbf{MG4}↑   & \textbf{MG5}↑   & \textbf{FID(C)↓} & \textbf{FID(F)↓} & \textbf{CLIP(C)↑} & \textbf{CLIP(F)↑} & \textbf{Latency}↓ \\
\midrule
SD v1.4 \cite{b40} & 0.2986 & 0.9072$_{1.33}$ & 0.5074$_{0.89}$ & 0.1148$_{0.45}$ & 0.0088$_{0.21}$ & 22.06 & 57.24 & 0.3022 & 0.3082 & \cellcolor{lightblue}7.54$_{0.17}$ \\
Composable \cite{b10} & 0.2783 & 0.6333$_{0.59}$ & 0.2187$_{1.01}$ & 0.0325$_{0.45}$ & 0.0023$_{0.18}$ & 21.71 & 60.32 & 0.3123 & 0.2987 & 13.81$_{0.15}$ \\
Layout \cite{b37} & 0.4359 & 0.9322$_{0.69}$ & 0.6015$_{1.58}$ & 0.1949$_{0.88}$ & 0.0227$_{0.44}$ & 23.18 & 58.77 & 0.2851 & 0.3042 & 18.89$_{0.20}$ \\
Structured \cite{b9} & 0.2964 & 0.9040$_{1.06}$ & 0.4864$_{1.32}$ & 0.1071$_{0.92}$ & 0.0068$_{0.25}$ & 22.09 & \cellcolor{lightgreen}56.73 & 0.2706 & 0.2931 & 7.74$_{0.17}$ \\
Attend-Excite \cite{b12} & 0.4513 & 0.9364$_{0.76}$ & 0.6510$_{1.24}$ & 0.2801$_{0.90}$ & \cellcolor{lightgreen}0.0601$_{0.61}$ & 21.57 & 57.05 & 0.3018 & 0.3044 & 25.43$_{4.89}$ \\

CoMat \cite{comat} & 0.4732 & 0.9241$_{0.33}$ & 0.7044$_{1.11}$ & 0.2566$_{0.77}$ & 0.0211$_{0.24}$ & 22.43 & 58.49 & 0.3122 & 0.3044 & 7.54$_{0.33}$ \\
IterComp \cite{itercomp} & 0.4891 & 0.9560$_{1.21}$ & 0.6742$_{1.54}$ & 0.2487$_{0.41}$ & 0.0167$_{0.15}$ & 21.68 & 57.66 & 0.3047 & 0.3152 & 7.54$_{0.35}$ \\
SynGen \cite{syngen} & 0.4570 & 0.9570$_{1.44}$ & 0.6915$_{0.68}$ & 0.2638$_{0.79}$ & 0.0241$_{0.46}$ & 22.35 & 56.94 & 0.3105 & 0.3201 & 9.21$_{0.26}$ \\
InitNO \cite{initnO3} & 0.3703 & 0.9144$_{0.67}$ & 0.6731$_{0.66}$ & 0.2544$_{0.30}$ & 0.0311$_{0.47}$ & 21.50 & 57.31 & 0.3120 & \cellcolor{lightgreen}0.3234 & 12.14$_{0.47}$ \\
Self-Cross \cite{self-cross} & 0.4461 & 0.9357$_{0.56}$ & 0.7144$_{1.68}$ & 0.2615$_{0.62}$ & 0.0262$_{0.38}$ & 23.11 & 57.26 & \cellcolor{lightgreen}0.3145 & 0.3154 & 10.35$_{0.53}$ \\

TokenCompose \cite{b8} & \cellcolor{lightgreen}0.5215 & \cellcolor{lightblue}0.9808$_{0.40}$ & \cellcolor{lightgreen}0.7616$_{1.04}$ & \cellcolor{lightgreen}0.2881$_{0.95}$ & 0.0328$_{0.48}$ & \cellcolor{lightgreen}21.12 & 56.93 & 0.3112 & 0.3205 & 7.56$_{0.14}$ \\
\hline
\textbf{ELDiff(Ours)} & \cellcolor{lightblue}0.5563 & \cellcolor{lightgreen}0.9755$_{0.65}$ & \cellcolor{lightblue}0.8041$_{1.15}$ & \cellcolor{lightblue}0.3305$_{1.58}$ & \cellcolor{lightblue}0.0931$_{0.47}$ & \cellcolor{lightblue}19.25 & \cellcolor{lightblue}55.13 & \cellcolor{lightblue}0.3276 & \cellcolor{lightblue}0.3359 & \cellcolor{lightblue}7.54$_{0.24}$ \\
\bottomrule
\end{tabular}}
\label{table1}
\end{table*}

\begin{table*}[t]
\caption{Comparison results with training-free methods on T2I-CompBench.}
\centering
\small
\setlength{\tabcolsep}{5pt}
\renewcommand{\arraystretch}{1.1}
\resizebox{0.75\textwidth}{!}{
\begin{tabular}{lccccccc}
\toprule
\multicolumn{1}{c}{\multirow{2}[2]{*}{\textbf{Method}}} & \multicolumn{3}{c}{\textbf{Attribute Binding}} & \multicolumn{2}{c}{\textbf{Object Relationship}} & \multicolumn{1}{c}{\multirow{2}[2]{*}{\textbf{Complex}↑}} & \multicolumn{1}{c}{\multirow{2}[2]{*}{\textbf{Latency}↓}}\\
\cmidrule(lr){2-4} \cmidrule(lr){5-6} 
& \textbf{Color}↑   & \textbf{Shape}↑  & \textbf{Texture}↑   & \textbf{Spatial}↑ & \textbf{Non-Spatial}↑ \\
\midrule
\rowcolor{rowgray} SDXL & 0.6734 & 0.5064 & 0.6243 & 0.2073 & 0.3166 & 0.3575 & 8.07s \\
SynGen \color{black}\cite{syngen} & \cellcolor{lightgreen}0.7267 & \cellcolor{lightgreen}0.5249 & 0.6476 & 0.2387 & \cellcolor{lightgreen}0.3172 & 0.3944 & 11.62s \\
InitNO \cite{initnO3} & 0.6823 & 0.5229 & 0.6547 & \cellcolor{lightblue}0.2553 & 0.3147 & \cellcolor{lightgreen}0.4077 & 19.42s\\
R2F \color{black}\cite{r2f}  & 0.7135 & 0.5194 & \cellcolor{lightgreen}0.6625 & 0.2447 & 0.3170 & 0.4031 & 10.14s \\
Ours & \cellcolor{lightblue}0.7768 & \cellcolor{lightblue}0.5663 & \cellcolor{lightblue}0.6941 & \cellcolor{lightgreen}0.2496 & \cellcolor{lightblue}0.3261 & \cellcolor{lightblue}0.4252 & \cellcolor{lightblue}8.07s\\
\midrule
\rowcolor{rowgray} SD v2.1 & 0.5309 & 0.4447 & 0.4929 & 0.1355 & 0.3099 &  0.3185 & 3.25s\\
Self-Cross \color{black}\cite{self-cross} & \cellcolor{lightgreen}0.6344 & 0.5347 & \cellcolor{lightgreen}0.5681 & \cellcolor{lightgreen}0.2048 & \cellcolor{lightgreen}0.3116 & \cellcolor{lightgreen}0.3664 & 13.05s \\
CountGui \color{black}\cite{CountGui} & 0.5387 & \cellcolor{lightblue}0.5371 & 0.5211 & 0.1339 & 0.3077 & 0.3296 & 11.74s \\
\textbf{Ours} & \cellcolor{lightblue}0.6604 & \cellcolor{lightgreen}0.5125 & \cellcolor{lightblue}0.6287 & \cellcolor{lightblue}0.2238 & \cellcolor{lightblue}0.3204 & \cellcolor{lightblue}0.3841 & \cellcolor{lightblue}3.25s\\
\bottomrule
\end{tabular}}
\label{table2}
\end{table*}

\section{Experiments}
\subsection{Datasets and Evaluation}
\textbf{Dataset for Finetuning.} We follow the dataset setting in TokenCompose \cite{b8}, which is a subset of COCO image-caption pairs \cite{b54}. To be specific, all unique images are from the Visual Spatial Reasoning dataset \cite{b55} since these image-caption pairs have relatively low ambiguity in its visual language and the rich  diversity of object categories. The CLIP model \cite{b35} is used to choose the caption that has the highest semantic correspondence to its paired image. The Grounded-SAM \cite{b15,b56} is utilized to generate binary segmentation maps of all nouns from the captions. Finally, 4526 image-caption pairs and their corresponding binary segmentation maps are finally obtained.

\noindent
\textbf{Evaluation metrics.} We measure the following metrics: VISOR \cite{b58}, MULTIGEN \cite{b8}, Fr$\acute{\textrm{e}}$chet Inception Distance (FID) \cite{b59}, CLIP Score \cite{b62}, Efficiency, T2I-CompBench \cite{b63}, T2I-CompBench++ \cite{T2i-compbench++}, GenEval \cite{geneval}, and GenEval 2 \cite{GenEval2}. The detailed evaluation dataset and calculation rules based on these metrics are described in \color{red}Appendix C\color{black}.

\subsection{Implementation Details and Baseline Methods}
\textbf{Implementation Details.} We follow the experiment settings outlined in TokenCompose \cite{b8}, finetuning our ELDiff based on Stable Diffusion v1.4 \cite{b29}. All experiments are implemented with PyTorch and carried out with NVIDIA GeForce RTX 3090 GPU. The number of training steps is 24000, the initial learning rate is 5e-6 with AdamW \cite{b61}, and the batch size is 1 and 4 gradient accumulation steps on a single GPU. The pseudocode of ELDiff is provided in \color{red}Appendix D\color{black}. Apart from the original denoised target $\mathcal{L}_{LDM}$, $\mathcal{L}_{PixEvi}$ and $\mathcal{L}_{\mathrm{PixEvi}}$ and $\mathcal{L}_{\mathrm{TokCon}}$ are utilized in the finetuning process, where $(\lambda_1,\lambda_2,\lambda_3)$ is set to (5e-5, 5e-7, 5e-7$\cdot$$min\{1,n_{epoch}/100\}$ for $\mathcal{L}_{\mathrm{PixEvi}}$ and $(\eta_1,\eta_2)$ is set to (1e-4, 1e-4) for $\mathcal{L}_{\mathrm{TokCon}}$. Hyperparameter ablation study results are provided in \color{red}Appendix E\color{black}. Binary masks are generated once offline using SAM before diffusion training, not in training process. No mask needed at testing. Training cost: one 3090 GPU, about 22 hours, comparable to Stable Diffusion.

\noindent
\textbf{Baseline Methods.} To validate the effectiveness of ELDiff, we conduct comparisons on multiple diffusion backbones, including SD v1.4 \cite{b40}, SD v2.1, SDXL \cite{sdxl}, SD v3.5 \cite{b64}, and Qwen-Image \cite{qwenimage}. On these backbones, we compare ELDiff with several representative T2I methods, including Composable Diffusion \cite{b10}, Layout Guidance Diffusion \cite{b37}, Structured Diffusion \cite{b9}, Attend-and-Excite \cite{b12}, CoMat \cite{comat}, IterComp \cite{itercomp}, SynGen \cite{syngen}, InitNO \cite{initnO3}, Self-Cross \cite{self-cross}, and TokenCompose \cite{b8}, R2F \cite{r2f}, and CountGui \cite{CountGui}.

\begin{table*}[t]
\caption{To evaluate the model generalization, we apply our finetuning strategy to SD v2.1 and SD v3.5 Medium, and report results on multi-category instance composition, FID, and CLIP score.}
\centering
\small
\setlength{\tabcolsep}{5pt}
\renewcommand{\arraystretch}{1.1}
\resizebox{0.85\textwidth}{!}{
\begin{tabular}{lcccccccc}
\toprule
\multicolumn{1}{c}{\multirow{2}[2]{*}{\textbf{Model}}} & \multicolumn{4}{c}{\textbf{Multi-category Instance Composition}} & \multicolumn{4}{c}{\textbf{Photorealism}} \\
\cmidrule(lr){2-5} \cmidrule(lr){6-9}
& \textbf{OA}↑   & \textbf{MG3}↑  & \textbf{MG4}↑   & \textbf{MG5}↑   & \textbf{FID(C)}↓ & \textbf{FID(F)}↓ & \textbf{CLIP(C)}↑ & \textbf{CLIP(F)}↑ \\
\midrule
SD v2.1 frozen & 0.4728 & 0.7014 & 0.2557 & 0.0327 & 21.79 & 56.83 & 0.3045 & 0.3159 \\
SD v2.1 ft. w. $\mathcal{L}_{LDM}$ & 0.5509 & 0.7643 & 0.3207 & 0.0473 & 21.94 & 57.02 & 0.3122 & 0.3231 \\

SD v2.1 ft. w. CoMat & 0.5844 & 0.7858 & 0.3279 & 0.0496 & 21.44 & 56.52 & 0.3168 & 0.3185 \\
SD v2.1 ft. w. IterComp & \cellcolor{lightgreen}0.6341 & \cellcolor{lightgreen}0.8072 & 0.3544 & 0.0543 & 20.85 & \cellcolor{lightgreen}56.31 & 0.3155 & 0.3237 \\

SD v2.1 ft. w. TokenCompose & 0.6010 & 0.8048 & \cellcolor{lightgreen}0.3669 & \cellcolor{lightgreen}0.0571 & \cellcolor{lightgreen}20.77 & 56.43 & \cellcolor{lightgreen}0.3209 & \cellcolor{lightgreen}0.3264 \\
SD v2.1 ft. w. Ours & \cellcolor{lightblue}0.7116 & \cellcolor{lightblue}0.8870 & \cellcolor{lightblue}0.5401 & \cellcolor{lightblue}0.1392 & \cellcolor{lightblue}20.09 & \cellcolor{lightblue}55.12 & \cellcolor{lightblue}0.3241 & 0.\cellcolor{lightblue}3348 \\
\hline
SD v3.5 frozen & 0.8084 & \cellcolor{lightgreen}0.9996 & 0.9696 & 0.7328 & 18.21 & \cellcolor{lightblue}54.33 & 0.3185 & 32.66 \\
SD v3.5 ft. w. $\mathcal{L}_{LDM}$ & 0.8010 & \cellcolor{lightgreen}0.9996 & 0.9702 & 0.7345 & 18.94 & 55.14 & 0.3174 & 32.43 \\

SD v3.5 ft. w. CoMat & 0.8122 & \cellcolor{lightblue}0.9997 & 0.9733 & 0.7456 & 18.45 & 55.01 & 0.3148 & \cellcolor{lightblue}32.93 \\
SD v3.5 ft. w. IterComp& 0.8208 & \cellcolor{lightblue}0.9997 & 0.9814 & \cellcolor{lightgreen}0.7644 & 18.21 & 54.85 & 0.3183 & 32.52 \\

SD v3.5 ft. w. TokenCompose & \cellcolor{lightgreen}0.8233 & \cellcolor{lightblue}0.9997 & \cellcolor{lightgreen}0.9839 & 0.7569 & \cellcolor{lightgreen}18.05 & 54.87 & \cellcolor{lightgreen}0.3198 & 32.87 \\
SD v3.5 ft. w. Ours & \cellcolor{lightblue}0.8644 & \cellcolor{lightblue}0.9997 & \cellcolor{lightblue}0.9964 & \cellcolor{lightblue}0.8171 & \cellcolor{lightblue}17.66 & \cellcolor{lightgreen}54.79 & \cellcolor{lightblue}0.3247 & \cellcolor{lightgreen}32.78 \\
\bottomrule
\end{tabular}}
\label{table3}
\end{table*}

\begin{table*}[t]
\caption{Evaluation results about compositionality on T2I-CompBench. Based on SD v2.1 and SD v3.5 Medium, ELDiff consistently shows the best performance regarding attribute binding, object relationships, numeracy and complex.}
\centering
\small
\setlength{\tabcolsep}{5pt}
\renewcommand{\arraystretch}{1.1}
\resizebox{0.8\textwidth}{!}{
\begin{tabular}{lcccccc}
\toprule
\multicolumn{1}{c}{\multirow{2}[2]{*}{\textbf{Model}}} & \multicolumn{3}{c}{\textbf{Attribute Binding}} & \multicolumn{2}{c}{\textbf{Object Relationship}} & \multicolumn{1}{c}{\multirow{2}[2]{*}{\textbf{Numeracy}↑}} \\
\cmidrule(lr){2-4} \cmidrule(lr){5-6}
& \textbf{Color}↑   & \textbf{Shape}↑  & \textbf{Texture}↑   & \textbf{Spatial}↑ & \textbf{Non-Spatial}↑ \\
\midrule
SD v2.1 frozen & 0.5309 & 0.4447 & 0.4929 & 0.1355 & 0.3099 & 0.4896 \\
SD v2.1 ft. w. $\mathcal{L}_{LDM}$ & 0.5514 & 0.5007 & 0.5376 & 0.1644 & \cellcolor{lightgreen}0.3192 & 0.5042 \\

SD v2.1 ft. w. CoMat & 0.6146 & 0.4835 & 0.5691 & 0.1543 & 0.3144 & 0.5078 \\
SD v2.1 ft. w. IterComp & \cellcolor{lightgreen}0.6354 & \cellcolor{lightgreen}0.5011 & 0.6124 & 0.1674 & 0.3169 & \cellcolor{lightgreen}0.5136 \\

SD v2.1 ft. w. TokenCompose & 0.6063 & 0.4934 & \cellcolor{lightgreen}0.6131 & \cellcolor{lightgreen}0.1789 & 0.3185 & 0.5094 \\
SD v2.1 ft. w. Ours & \cellcolor{lightblue}0.6604 & \cellcolor{lightblue}0.5325 & \cellcolor{lightblue}0.6487 & \cellcolor{lightblue}0.2238 & \cellcolor{lightblue}0.3204 & \cellcolor{lightblue}0.5366 \\
\hline
SD v3.5 frozen & 0.8033 & 0.5831 & 0.7154 & 0.3102 & 0.4010 & 0.6043 \\
SD v3.5 ft. w. $\mathcal{L}_{LDM}$ & 0.8152 & 0.5641 & 0.7184 & 0.3361 & 0.4122 & 0.5961 \\

SD v3.5 ft. w. CoMat & 0.8124 & 0.5914 & 0.7283 & \cellcolor{lightgreen}0.3681 & 0.4451 & 0.6037 \\
SD v3.5 ft. w. IterComp & \cellcolor{lightgreen}0.8177 & \cellcolor{lightgreen}0.6035 & 0.7358 & 0.3422 & 0.4295 & \cellcolor{lightgreen}0.6254 \\

SD v3.5 ft. w. TokenCompose & 0.8023 & 0.5748 & \cellcolor{lightblue}0.7463 & 0.3651 & \cellcolor{lightgreen}0.4474 & 0.6243 \\
SD v3.5 ft. w. Ours & \cellcolor{lightblue}0.8452 & \cellcolor{lightblue}0.6241 & \cellcolor{lightgreen}0.7435 & \cellcolor{lightblue}0.3925 & \cellcolor{lightblue}0.4640 & \cellcolor{lightblue}0.6613 \\
\bottomrule
\end{tabular}}
\label{table4}
\end{table*}

\begin{table*}[t]
\caption{Ablation studies of different objectives with $\mathcal{L}_{LDM}$, $\mathcal{L}_{\mathrm{PixEvi}}$, and $\mathcal{L}_{\mathrm{TokCon}}$.}
\centering
\small
\setlength{\tabcolsep}{5pt}
\renewcommand{\arraystretch}{1.1}
\resizebox{\textwidth}{!}{
\begin{tabular}{ccccccccccccc}
\toprule
\multirow{2}[2]{*}{\textbf{Model}} & \multirow{2}[2]{*}{$\mathcal{L}_{LDM}$} & \multirow{2}[2]{*}{$\mathcal{L}_{\mathrm{PixEvi}}$} & \multirow{2}[2]{*}{$\mathcal{L}_{\mathrm{TokCon}}$} & \multicolumn{5}{c}{\textbf{Multi-category Instance Composition}} & \multicolumn{4}{c}{\textbf{Photorealism}} \\
\cmidrule(lr){5-9} \cmidrule(lr){10-13}
&       &       &       & \textbf{OA}↑    & \textbf{MG2}↑   & \textbf{MG3}↑   & \textbf{MG4}↑   & \textbf{MG5}↑   & \textbf{FID(C)}↓ & \textbf{FID(F)}↓ & \textbf{CLIP(C)}↑ & \textbf{CLIP(F)}↑ \\
\midrule
SD v1.4 &       &       &       & 0.2986 & 0.9072$_{1.33}$ & 0.5074$_{0.89}$ & 0.1148$_{0.45}$ & 0.0088$_{0.21}$ & 22.06 & 57.24 & 0.3022 & 0.3082 \\
SD v1.4 & \checkmark   &       &       & 0.3821 & 0.9144$_{0.21}$ & 0.6372$_{1.11}$ & 0.1956$_{0.94}$ & 0.1897$_{0.32}$ & 24.57 & 60.14 & 0.3028 & 0.3106 \\
SD v1.4 & \checkmark   & \checkmark   &       & \cellcolor{lightgreen}0.5466 & \cellcolor{lightblue}0.9821$_{0.30}$ & \cellcolor{lightgreen}0.7648$_{0.88}$ & \cellcolor{lightgreen}0.2934$_{1.14}$ & \cellcolor{lightgreen}0.3361$_{0.21}$ & \cellcolor{lightgreen}21.17 & \cellcolor{lightgreen}55.98 & \cellcolor{lightgreen}0.3178 & \cellcolor{lightgreen}0.3246 \\
SD v1.4 & \checkmark   & \checkmark   & \checkmark   & \cellcolor{lightblue}0.5563 & \cellcolor{lightgreen}0.9736$_{0.65}$ & \cellcolor{lightblue}0.7660$_{1.15}$ & \cellcolor{lightblue}0.2962$_{1.58}$ & \cellcolor{lightblue}0.3530$_{0.47}$ & \cellcolor{lightblue}20.84 & \cellcolor{lightblue}55.36 & \cellcolor{lightblue}0.3204 & \cellcolor{lightblue}0.3299 \\
\bottomrule
\end{tabular}}
\label{table5}
\end{table*}

\begin{table*}[t]
\caption{Comparison results on GenEval using the stronger Qwen-Image backbone.}
\centering
\small
\setlength{\tabcolsep}{5pt}
\renewcommand{\arraystretch}{1.1}
\resizebox{\textwidth}{!}{
\begin{tabular}{l|cccccc|c}
\toprule
\textbf{Model} &
\textbf{Single Object} &
\textbf{Two Object} &
\textbf{Counting} &
\textbf{Colors} &
\textbf{Position} &
\textbf{Attribute Binding} &
\textbf{Overall} \\
\midrule
Qwen-Image \cite{qwenimage} & \cellcolor{lightgreen}0.99 & 0.92 & \cellcolor{lightgreen}0.89 & 0.88 & 0.76 & \cellcolor{lightgreen}0.77 & 0.87 \\
ft. w. TokenCompose & \cellcolor{lightblue}1.00 & \cellcolor{lightgreen}0.93 & \cellcolor{lightgreen}0.89 & \cellcolor{lightgreen}0.89 & \cellcolor{lightgreen}0.77 & \cellcolor{lightblue}0.79 & \cellcolor{lightgreen}0.88 \\
ft. w. Ours & \cellcolor{lightblue}1.00 & \cellcolor{lightblue}0.96 & \cellcolor{lightblue}0.90 & \cellcolor{lightblue}0.92 & \cellcolor{lightblue}0.80 & \cellcolor{lightblue}0.79 & \cellcolor{lightblue}0.90 \\

\bottomrule
\end{tabular}}
\label{table6}
\end{table*}

\begin{table}[t]
\centering
\caption{Comparison of Qwen-Image on GenEval 2.}
\resizebox{0.75\textwidth}{!}{
\begin{tabular}{l|ccccc|cc}
\toprule
Methods & Attribute$\uparrow$ & Count$\uparrow$ & Object$\uparrow$ & Position$\uparrow$ & Verb$\uparrow$ & Soft-TIFA$_{\mathrm{AM}}\uparrow$ & Soft-TIFA$_{\mathrm{GM}}\uparrow$ \\
\midrule
Qwen-Image    & 84.11 & 67.98 & 99.30 & 60.82 & 37.50 & 81.58 & 34.57 \\
ft. w. $\mathcal{L}_{LDM}$  & 84.33 & 68.47 & 99.46 & 60.93  & 38.07 & 81.90 & 35.11   \\
ft. w. TokenCom & 84.42 & 68.55 & 99.40 & 61.54 & 37.88 & 81.79 & 35.83 \\
ft. w. Ours          & \cellcolor{lightblue}85.41 & \cellcolor{lightblue}70.67 & \cellcolor{lightblue}99.58 & \cellcolor{lightblue}61.97 & \cellcolor{lightblue}39.40 & \cellcolor{lightblue}83.13 & \cellcolor{lightblue}37.94 \\
\bottomrule
\end{tabular}
}
\label{table7}
\end{table}

\begin{table}[t]
\centering
\caption{Comparison of Qwen-Image on T2I-CompBench++.}
\resizebox{0.75\textwidth}{!}{
\begin{tabular}{l|cccccccc}
\toprule
Methods & Color$\uparrow$ & Shape$\uparrow$ & Texture$\uparrow$ & 2D Spatial$\uparrow$ & 3D Spatial$\uparrow$ & Non-Spatial$\uparrow$ & Numeracy$\uparrow$ & Complex$\uparrow$ \\
\midrule
Qwen-Image    & 83.92 & 59.41 & 75.68 & 45.12 & 46.31 & 31.26 & 77.05 & 39.44 \\
ft. w. $\mathcal{L}_{LDM}$ & 84.74 & 59.48 & 75.83 & 45.69 & 46.44 & 31.34 & 77.40 & 39.50 \\
ft. w. TokenCom & 84.44 & 59.57 & \cellcolor{lightblue}76.41 & 46.37 & 46.86 & 31.31 & 77.64 & 39.62 \\
ft. w. Ours          & \cellcolor{lightblue}85.25 & \cellcolor{lightblue}61.67 & 76.17 & \cellcolor{lightblue}48.55 & \cellcolor{lightblue}47.61 & \cellcolor{lightblue}31.53 & \cellcolor{lightblue}78.12 & \cellcolor{lightblue}40.24 \\
\bottomrule
\end{tabular}
}
\label{table8}
\end{table}

\begin{table}[h]
\caption{Results of user study on SD v2.1 and SD v3.5.}
\centering
\small
\setlength{\tabcolsep}{5pt}
\renewcommand{\arraystretch}{1.1}
\resizebox{0.7\textwidth}{!}{
\begin{tabular}{lccc}
\toprule
\textbf{Method}
& \textbf{Image Realism}↑ 
& \textbf{Image Compositionality}↑ 
& \textbf{Overall}↑ \\
\midrule
SD v2.1 & 12.2\% & 5.8\% & 19.7\% \\
SD v2.1 ft. w. TokenCompose & \cellcolor{lightgreen}34.3\% & \cellcolor{lightgreen}36.8\% & \cellcolor{lightgreen}31.7\% \\
SD v2.1 ft. w. Ours & \cellcolor{lightblue}53.5\% & \cellcolor{lightblue}57.4\% & \cellcolor{lightblue}48.6\% \\
\midrule
SD v3.5 & \cellcolor{lightgreen}30.1\% & 25.0\% & 29.6\% \\
SD v3.5 ft. w. TokenCompose & 27.4\% & \cellcolor{lightgreen}35.7\% & \cellcolor{lightgreen}31.5\% \\
SD v3.5 ft. w. Ours & \cellcolor{lightblue}42.5\% & \cellcolor{lightblue}42.6\% & \cellcolor{lightblue}38.9\% \\
\bottomrule
\end{tabular}}
\label{table9}
\end{table}

\begin{figure}[t]
\centering
\includegraphics[width=0.5\linewidth]{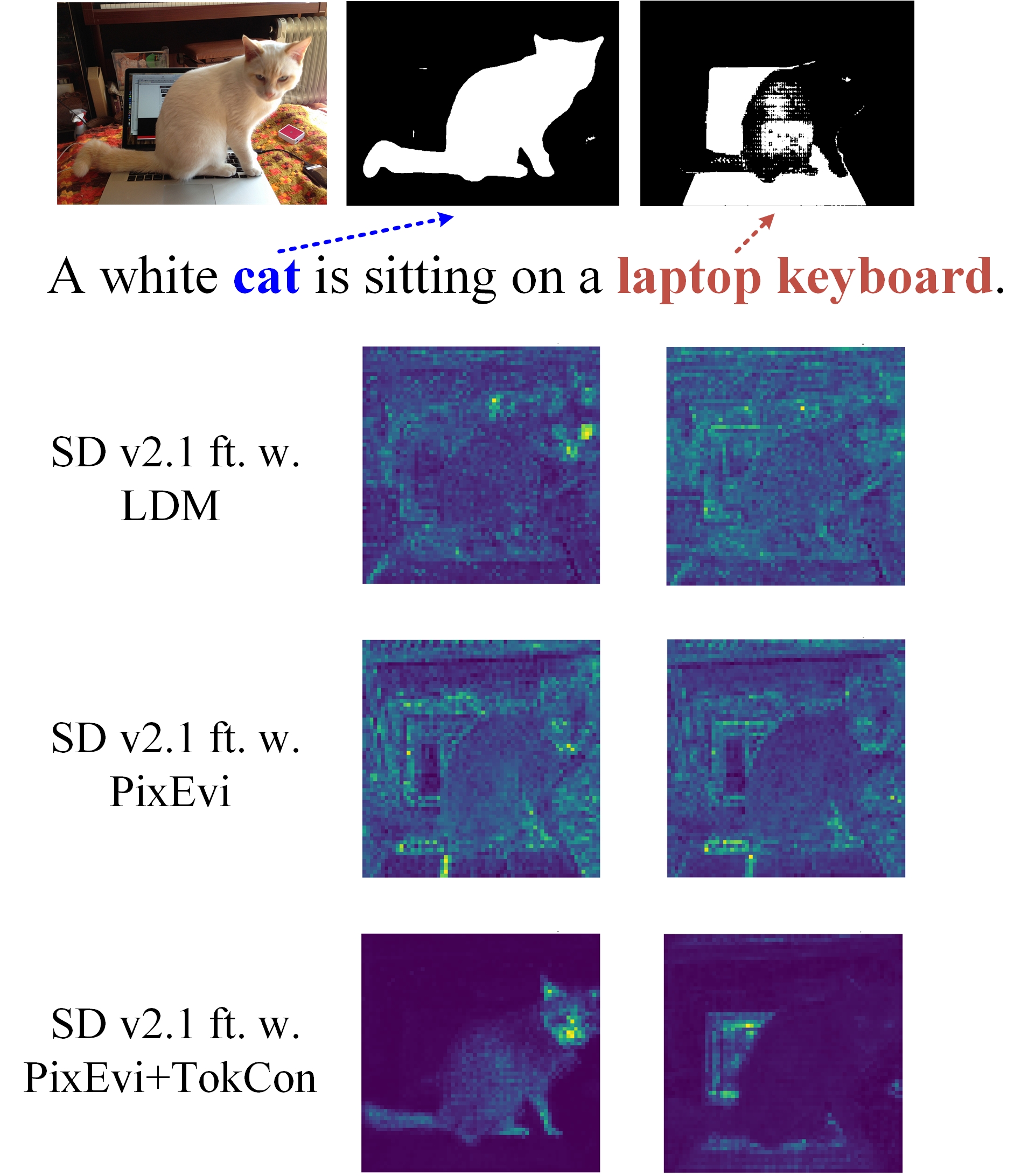}
\caption{Ablation visualization demonstrates that finetuning the Stable Diffusion with only $\mathcal{L}_{LDM}$ does not clearly identify the target objects. By introducing $\mathcal{L}_{PixEvi}$ and $\mathcal{L}_{TokCon}$, the model shows substantial improvement in text-object correspondence.}
\label{fig4}
\end{figure}

\begin{figure*}[t]
\centering
\includegraphics[width=\linewidth]{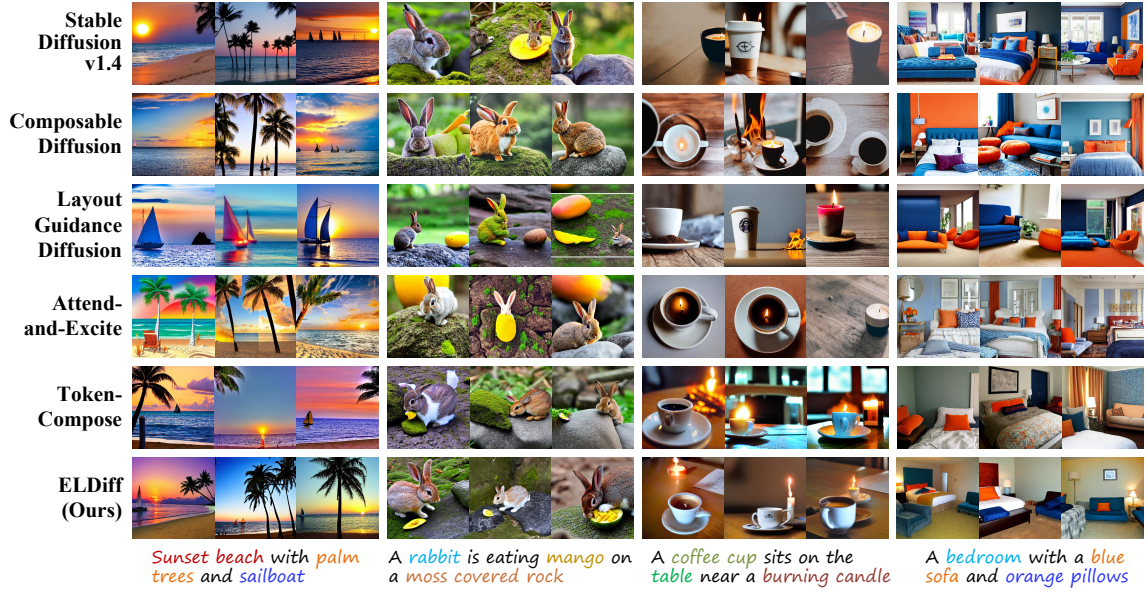}
\caption{Qualitative comparison between our ELDiff and the other T2I models.}
\label{fig5}
\end{figure*}

\subsection{Main Results}
\textbf{Results of Multi-category Instance Composition: Object Accuracy (OA) \& MULTIGEN} $\quad$ We quantitatively evaluate the compositionality of ELDiff based on OA (a metric from VISOR) \cite{b58} and MULTIGEN \cite{b8} compared to the outstanding T2I models. As demonstrated in Table \ref{table1}, ELDiff achieves state-of-the-art performance on OA, which is 0.0348 higher than the suboptimal TokenCompose \cite{b8}. For MG2-5, it is clear that our ELDiff and TokenCompose show significant improvements apart from MG2. To verify whether our method is benefit for finetuning other SD versions, we add our fine-tuning strategy to Stable Diffusion v2.1 and SD v3.5 Medium. As shown in Table \ref{table3}, the results of Multi-category Instance Composition and Photorealism indicate that our ELDiff significantly outperforms SD v2.1, SD v3.5, and other finetuning methods. On SD v2.1, compared to TokenCompose, our ELDiff improved OA by 0.1106 and MG4 by 0.1732. In addition, on SD v3.5, our method has also achieves competitive performance. These improvements are largely attributed to the consistency constraint between the text prompts and the image contents, which greatly enhances the model’s ability of composing multiple object categories. By finetuning the Stable Diffusion through pixel evidence loss and token conflict loss, ELDiff provides reliable consistency constraint during the denoising process and achieves outstanding generation results involving multiple objects. For qualitative experiments, as shown in Fig. \ref{fig5}, ELDiff achieves a high level of image quality in conjunction with multi-category instance composition, compared with other T2I models. The generation results of different finetuning methods based on SD v2.1 and SD v3.5 are provided in \color{red}Appendix F\color{black}.

\noindent
\textbf{Results of Photorealism: FID \& CLIP Score} $\quad$ As shown in Table \ref{table1}, our ELDiff consistently outperforms other T2I models in both FID \cite{b59} and CLIP Score \cite{b62}. Specifically, ELDiff is 1.87 and 1.60 lower than the suboptimal method on Fid(C) and Fid(F), respectively. It also 0.0131 and 0.0125 higher than the suboptimal method on CLIP Score(C) and CLIP Score(F) respectively. We attribute these performance advantages to the proposed pixel evidence loss and token conflict loss, which enhances image photorealism when composing multiple categories of instances. 

\noindent
\textbf{Results of Efficiency: Latency} $\quad$ Compared to the standard T2I diffusion, our ELDiff does not require additional inference time during the inference stage. As shown in Table \ref{table1}, ELDiff achieves the best generation performance using the least amount of inference time. The latency results in Table \ref{table1} represent the number of seconds required to generate an image with 50 DDIM steps.

\noindent
\textbf{Results of Attribute Binding and Object Relationship: T2I-CompBench} $\quad$ As illustrated in the Table \ref{table4}, our ELDiff observes significant gains in all seven evaluation tasks, compared with other finetuing methods. Based on SD v2.1, ELDiff outperforms the second-best method by 0.025, 0.0314, 0.0356, 0.0449, and 0.0230 on color, shape, texture, spatial, and numeracy. On SD3.5, ELDiff also consistently outperforms the compared approaches. It is notable that ELDiff improves attribute binding and object relationship by enhancing the model's multi-object composition ability without specifically optimizing color, shape, texture, and the relationships between objects.

\subsection{Ablation Study}
\textbf{Importance of Pixel Evidence Loss $\mathcal{L}_{\mathrm{PixEvi}}$} $\quad$ As shown in Table \ref{table5}, we show the Multi-category Instance Composition and Photorealism results, aiming to identify the effectiveness of pixel evidence loss. It is clear that without the use of the pixel evidence loss, Multi-category Instance Composition and Photorealism  have significantly worsened. This is because the text prompt and the object in the image are not aligned during the denoising process, leading to poor multi-categories generation.

\noindent
\textbf{Effectiveness of Token Conflict Loss $\mathcal{L}_{\mathrm{TokCon}}$} $\quad$ We demonstrate the advantages of token conflict loss by introducing the token conflict loss into the pixel evidence loss. As shown in Table \ref{table5}, compared to before adding the token conflict loss, the performance metrics of Multi-category Instance Composition and Photorealism after adding the token conflict loss achieves consistent improvements apart from MG2. This is because the token conflict loss avoids multi-object semantic overlap conflicts in the image when conducting the consistency constraint between text prompts and image contents.

The ablation visualization (Fig. \ref{fig4}) clearly shows that finetuning Stable Diffusion with only $\mathcal{L}_{LDM}$ struggles to accurately capture the target objects. When the pixel evidence loss $\mathcal{L}_{PixEvi}$ is incorporated, the model improves the reliability of object localization. Furthermore, adding the token conflict loss $\mathcal{L}_{TokCon}$ results in a enhancement of text-to-object consistency. The ablation visualization demonstrates the effectiveness of the proposed loss in capturing the objects corresponding to nouns in the text.

\subsection{Comparison with Training-Free Methods}
Table \ref{table2} reports the comparison with representative training-free methods on T2I-CompBench under SDXL and SD v2.1 backbones. Our ELDiff achieves the best overall performance across most metrics. On SDXL, compared to the second-best method, ELDiff improves attribute binding (color, shape, and texture) by 0.0504, 0.0414, and 0.0316, and obtains the highest complex score while maintaining competitive latency. Similar trends are observed on SD v2.1, where our method consistently outperforms Self-Cross and CountGui in both attribute binding and spatial relationships.

\subsection{Comparison with Stronger T2I Backbone}
Table \ref{table6} shows the GenEval results on the stronger Qwen-Image backbone. Specifically, we integrate both our method and the TokenCompose finetuning strategy into Qwen-Image for comparison. Our method consistently outperforms Qwen-Image and TokenCompose. The overall score increases from 0.87 to 0.90. In addition, Table \ref{table7} shows that our finetuning method consistently outperforms Qwen-Image on GenEval 2, with gains of 1.55 and 3.37 on Soft-TIFA$_{\mathrm{AM}}$ and Soft-TIFA$_{\mathrm{GM}}$. We used Qwen3-VL-8B \cite{Qwen3-vl} as the VLM evaluator. On T2I-ComBench++ (Table \ref{table8}), compared with Qwen-Image, it improves Color, Shape, Texture, 2D Spatial, 3D Spatial, Non-Spatial, Numeracy, and Complex scores by 1.33, 2.26, 0.49, 3.43, 1.30, 0.27, 1.07, and 0.80, respectively. Additionally, compared with SFT loss ($\mathcal{L}_{LDM}$), our method also shows consistent improvements. This demonstrates that our ELDiff has strong generalization capability on the stronger backbone.

\subsection{User Study}
We also conducted user studies on SD v2.1 and SD v3.5. This study involves 11 volunteers with expertise in image processing. Participants were asked to select the best image based on realism, compositionality, and overall quality. We use 1,000 prompts from DSG1K \cite{DSG1K} for evaluation. Voting results are shown in Table \ref{table9}. Across two backbones, ELDiff consistently ranked first in user evaluations.

\section{Conclusion}
In this paper, we formulate the problems of segmentation map bias and semantic overlap conflict in performing the consistency constraint between text prompts and image contents, and propose ELDiff, an end-to-end T2I diffusion model fine-tuning strategy equipped with consistency constraint between text prompts and image contents. We identify the causes of these two problems and propose two key components to explicitly address them. The pixel evidence loss judges the reliability of consistency supervision through pixel-level uncertainty metric. Besides, we introduce the token conflict loss to address the semantic overlap conflict among objects in the image. Despite achieving superior generation results, ELDiff shortcomings in attribute binding task. In future work, we will continue to improve this framework by considering applying constraints on color, shape, texture, etc.

\section*{Acknowledgements}
This work was partly supported by Taishan Scholars Program of Shandong Province, the National Natural Science Foundation of China (Grant No. 62173212), and Shandong Province ``Double Hundred Talent Plan'' on 100 Foreign Experts and 100 Foreign Expert Teams (Grant No.WSR2023049).

%
%
\bibliographystyle{splncs04}
\bibliography{main}

\newpage
\section*{Appendix}
This supplementary material contains several sections that provide additional details related to our work on ELDiff. Specifically, it will cover the following topics:

\begin{itemize}
    \item In Appendix \ref{appA}, we provide a preliminary of Stable Diffusion Model and Cross-attention layer.
    
    \item In Appendix \ref{appB.1}, we derivate the problem of segmentation map bias.

    \item In Appendix \ref{appB.2}, we derivate the problem of semantic overlap conflict.

    \item In Appendix \ref{appC}, we provide a detailed explanation of the evaluation metrics used in this work.

    \item In Appendix \ref{appD}, we provide the pseudocode for ELDiff to thoroughly demonstrate its finetuning and inferencing process.

    \item In Appendix \ref{appE}, we provide the hyperparameter ablation study results.

    \item In Appendix \ref{appF}, we provide the generation results of different finetuning methods based on Stable Diffusion v2.1 and SD v3.5 Medium.
\end{itemize}

\appendix
\section{Preliminary}
\label{appA}
\textbf{Stable Diffusion Model}. This work focuses on the Stable Diffusion Model (SD) [\citenum{b29}] which functions within the latent space of an autoencoder. To be specific, an image $x_0$ is encoded to a latent code $z_0=\mathcal{E}(x_0)$ using a variational autoencoder (VAE) [\citenum{b50}]. Subsequently, a normally distributed noise $\epsilon$ is added to the original latent code $z_0$ with a variable extent based on a timestep $t$ sampling from $\{1,...,T\}$. During denoising, the denoising function $\epsilon_{\theta}$ parameterized by a UNet [\citenum{b51}] backbone is trained to predict the noise added to $z_0$ with the text prompt $y$ and the current latent $z_t$ as the input. The text prompt $y$ is first encoded to the text embedding $c=f_\mathrm{CLIP}(y)$ through a pre-trained CLIP [\citenum{b35}] text encoder $f_\mathrm{CLIP}$. The following shows the denoising objective.
\begin{equation}
\mathcal{L}_{\mathrm{LDM}}=\mathbb{E}_{z_0,t,c,\epsilon\sim\mathcal{N}(0,I)}\left[\left\|\epsilon-\epsilon_\theta\left(z_t,t,c\right)\right\|^2\right]
\end{equation}

During inference, a latent variable $z_T$ is sampled from the standard Gaussian distribution $\mathcal{N}(0,1)$, and then iteratively uses $\epsilon_{\theta}$ to estimate the noise and compute the next latent sample, thus deriving a refined latent representation $z_0$.

\noindent
\textbf{Cross-attention layer}. Text image interaction in SD is achieved through the cross-attention layer, facilitating text condition guidance. Specifically, for each cross-attention layer, linear projections are employed to extract the key $K$ and value $V$ from $c$, while the query $Q$ is projected by the intermediate features of UNet. The cross-attention
map $A^{(k)}\in\mathbb{R}^{h\times w\times l}$ is computed by:
\begin{equation}
A^{(k)}=\mathrm{Softmax}(\frac{Q^{(k)}(K^{(k)})^{T}}{\sqrt{d}})
\end{equation}
where $k$ is the index of head. $h$ and $w$ are the resolution of the latent code. $l$ is the token length of the text embedding, and d is the feature dimension. $A_{i,j}^{n}$ is the attention score assigned to $n$-th text token for the $(x,y)$-th spatial patch of the intermediate feature map.

\section{Problem Derivation}
\subsection{Segmentation Map Bias}
\label{appB.1}
Assuming the cross-attention map is $A_{i,j}\in[0,1]$, it represents the probability that pixel $(i,j)$ belongs to the foreground. The true segmentation map is $M^*_{i,j}\in\{0,1\}$. The segmentation map actually provided for model training is $M_{i,j}\in\{0,1\}$, which is biased, defined as:
\begin{equation}
\varepsilon_{i,j}:=M_{i,j}-M^*_{i,j}\in\{-1,0,+1\}
\end{equation}
The model fits the segmentation map $M_{i,j}$ with the cross-attention map $A_{i,j}$. The loss is defined as follows:
\begin{equation}
\mathcal{L}=-\sum_{i,j}\left[M_{i,j}\cdot\log A_{i,j}+(1-M_{i,j})\cdot\log(1-A_{i,j})\right]
\end{equation}
Bring in $M_{i,j}=M^*_{i,j}+\varepsilon_{i,j}$:
\begin{equation}
\begin{split}
\mathcal{L} = -\sum_{i,j} \Big[
    (M^*_{i,j}+\varepsilon_{i,j})\log A_{i,j} \\
    + (1-M^*_{i,j}-\varepsilon_{i,j})\log(1-A_{i,j})
\Big]
\end{split}
\end{equation}
Unfold it as:
\begin{equation}
\begin{aligned}
\mathcal{L}
&= 
\underbrace{
-\sum_{i,j} \left[
M^*_{i,j}\log A_{i,j}
+ (1-M^*_{i,j})\log(1-A_{i,j})
\right]
}_{\text{Ideal supervision objective}}
\\[0.6em]
&\qquad
\underbrace{
-\sum_{i,j} \left[
\varepsilon_{i,j}\log\frac{A_{i,j}}{1-A_{i,j}}
\right]
}_{\text{Bias term}}
\end{aligned}
\end{equation}

For the Bias term:
\begin{equation}
\mathcal{L}_\mathrm{bias}:=-\sum_{i,j}\varepsilon_{i,j}\log\frac{A_{i,j}}{1-A_{i,j}}
\end{equation}
It represents the perturbation of the biased segmentation map for model learning. We analyze different situations as follows:

\textit{Situation 1: False Positive.} That is, $M^*_{i,j}=0$, $M_{i,j}=1$, then $\varepsilon_{i,j}=1$. The bias term becomes: $-\log\frac{A_{i,j}}{1-A_{i,j}}$. When $A_{i,j}<0.5$, the gradient is positive, and the decrease of loss drives $A_{i,j}$ to increase. Consequently, the model is encouraged to improve the response of the error area, incorrectly focusing on foreground area.

\textit{Situation 2: False Negative.}
That is, $M^*_{i,j}=1$, $M_{i,j}=0$, then $\varepsilon_{i,j}=-1$. The bias term becomes: $\log\frac{A_{i,j}}{1-A_{i,j}}$. When $A_{i,j}>0.5$, the gradient is positive, and the decrease of loss drives $A_{i,j}$ to decrease. Consequently, the model suppresses real foreground response, ignoring the real foreground area.

Gradient Analysis: For $A_{i,j}$, its gradient is:
\begin{equation}
\frac{\partial\mathcal{L}}{\partial A_{i,j}}=\frac{A_{i,j}-M_{i,j}}{A_{i,j}(1-A_{i,j})}
\end{equation}
In the ideal situation, it should be:
\begin{equation}
\frac{\partial\mathcal{L}_{\mathrm{ideal}}}{\partial A_{i,j}}=\frac{A_{i,j}-M^*_{i,j}}{A_{i,j}(1-A_{i,j})}
\end{equation}
Therefore, in actual training, the direction of model optimization is:
\begin{equation}
\frac{\partial\mathcal{L}}{\partial A_{i,j}}=\frac{A_{i,j}-M^*_{i,j}-\varepsilon_{i,j}}{A_{i,j}(1-A_{i,j})}=\frac{\partial\mathcal{L}_{\mathrm{ideal}}}{\partial A_{i,j}}-\frac{\varepsilon_{i,j}}{A_{i,j}(1-A_{i,j})}
\end{equation}
The bias term term $\frac{\varepsilon_{i,j}}{A_{i,j}(1-A_{i,j})}$ explicitly change the direction of the gradient, which makes the biased pixel dominate the optimization direction of the model, causing the model to fit $M^{*}+\varepsilon$ instead of the true segmentation map $M^{*}$.

\subsection{Semantic Overlap Conflict}
\label{appB.2}
Assuming the text prompt contains $N$ nouns $\{w_1,\ldots,w_N\}$, each noun $w_n$ has a corresponding true segmentation map $M^n\in\{0,1\}$. The model predicts the cross-attention map $A^n_{i,j}\in[0,1]$ for each $w_n$, defined as:
\begin{equation}
A^n_{i,j}=\sigma(\phi(z_{i,j},e(w_n)))\quad\textrm{for each pixel $(i,j)$}
\end{equation}
where $z_{i,j}\in\mathbb{R}^d$ is the visual feature of pixel $(i,j)$, $e(w_n)\in\mathbb{R}^d$ is the noun embedding, $\phi(\cdot)$ is the cross-attention fusion function, and $\sigma$ is the activation function.

Let the target areas of two nouns $w_1,w_2$ overlap, i.e., there exists a pixel $(i,j)$ such that:
\begin{equation}
M^1_{i,j}=1,\quad M^2_{i,j}=1
\end{equation}
We investigate whether the gradient direction of the shared image feature $z_{i,j}$ at the pixel $(i,j)$ is consistent, that is
\begin{equation}
\nabla_{z_{i,j}}\mathcal{L}_1\quad\mathrm{vs}\quad\nabla_{z_{i,j}}\mathcal{L}_2
\end{equation}
If they have significant differences, it indicates that there is a optimization conflict during the training process.

For noun $w_1$, Let:
\begin{equation}
a_1=\phi(z_{i,j},e(w_1)),\quad A^1_{i,j}=\sigma(a_1)
\end{equation}

Then:
\begin{equation}
\frac{\partial\mathcal{L}_1}{\partial z_{i,j}}=\frac{\partial\mathcal{L}_1}{\partial A^1_{i,j}}\cdot\frac{d A^1_{i,j}}{da_1}\cdot\frac{\partial a_1}{\partial z_{i,j}}
\end{equation}

Expand item by item:
\begin{equation}
\frac{\partial\mathcal{L}_1}{\partial A^1_{i,j}}=-\frac{M^1_{i,j}}{A^1_{i,j}}+\frac{1-M^1_{i,j}}{1-A^1_{i,j}}
\end{equation}
\begin{equation}
\frac{d A^1_{i,j}}{da_1}=A^1_{i,j}(1-A^1_{i,j})
\end{equation}
\begin{equation}
\textrm{if }\phi(z_{i,j},e(w_1))=z_{i,j}^\top e(w_1) \quad \textrm{then }\frac{\partial a_1}{\partial z_{i,j}}=e(w_1)
\end{equation}

Based on the above analysis:
\begin{equation}
\nabla_{z_{i,j}}\mathcal{L}_1=\left(-\frac{M^1_{i,j}}{A^1_{i,j}}+\frac{1-M^1_{i,j}}{1-A^1_{i,j}}\right)\cdot A^1_{i,j}(1-A^1_{i,j})\cdot e(w_1)
\end{equation}

By analogy:
\begin{equation}
\nabla_{z_{i,j}}\mathcal{L}_2=\left(-\frac{M^2_{i,j}}{A^2_{i,j}}+\frac{1-M^2_{i,j}}{1-A^2_{i,j}}\right)\cdot A^2_{i,j}(1-A^2_{i,j})\cdot e(w_2)
\end{equation}

Define the overlapping area between $w_1$ and $w_2$ as:
\begin{equation}
\mathcal{O}_{i,j}=\{(i,j)\in[H]\times[W]\mid M^1_{i,j}=1\mathrm{and}M^2_{i,j}=1\}
\end{equation}

For the pixel $(i,j)\in\mathcal{O}_{i,j}$, 
\begin{equation}
\begin{aligned}
\nabla_{z_{i,j}}\mathcal{L}_1
&\propto 
\left(-\frac{1}{A^1_{i,j}}\right)
A^1_{i,j}(1-A^1_{i,j}) \, e(w_1) \\[0.3em]
&= -(1-A^1_{i,j})\, e(w_1)
\end{aligned}
\end{equation}

By analogy:
\begin{equation}
\nabla_{z_{i,j}}\mathcal{L}_2\propto-(1-A^2_{i,j})\cdot e(w_2)
\end{equation}

Finally, we obtained:
\begin{equation}
\begin{aligned}
\nabla_{z_{i,j}}\mathcal{L}_{\mathrm{total}}
&= \nabla_{z_{i,j}}\mathcal{L}_1 + \nabla_{z_{i,j}}\mathcal{L}_2 \\[0.3em]
&= -(1-A^1_{i,j})\,e(w_1) \;-\; (1-A^2_{i,j})\,e(w_2) \\[0.3em]
&= -\big[(1-A^1_{i,j})\,e(w_1) + (1-A^2_{i,j})\,e(w_2)\big]
\end{aligned}
\end{equation}

This is a weighted sum of two nouns $w_1, w_2$. If $e(w_1)$ and $e(w_2)$ has semantic overlap, then there will be conflict in the overall gradient direction during optimization.

Specifically, when:
\begin{equation}
\langle e(w_1),e(w_2)\rangle<0
\end{equation}

The two gradient directions cancel each other out, ultimately leading to:
\begin{equation}
\|\nabla_{z_{i,j}}\mathcal{L}_{\mathrm{total}}\|<\min\{\|\nabla_{z_{i,j}}\mathcal{L}_1\|,\|\nabla_{z_{i,j}}\mathcal{L}_2\|\}
\end{equation}
It causes the gradient conflict.

\section{Details of Evaluation metrics}
\label{appC}
We measure the following metrics: \textbf{1) VISOR} [\citenum{b58}] can assess how accurately the spatial relationship described in text is generated in the image. The VISOR benchmark generates all unique pairwise combinations of 80 COCO object categories. Each pair (A, B) is converted into a text prompt following a template “<A><R><B>,” where R represents an arbitrary spatial relationship. For example, one such prompt could be “a cat to the right of a table.” Finally, 31600 text prompts were constructed and used as conditions to generate 31600 images. Similarity, the open-vocabulary detector [\citenum{b57}] is used to detect the presence of each category in each generated image and the Object Accuracy (OA) is employed to evaluate how successfully instances are generated from each of the two categories. \textbf{2) MULTIGEN} [\citenum{b8}] is used to evaluate the ability of the model in combining instances of multiple categories. Specifically, given a set of $N$ distinct instance categories, five categories (e.g., A, B, C, D, and E) are randomly selected and formatted into a sentence (e.g., "A photo of A, B, C, D, and E"). This sentence serves as the conditional input for the T2I diffusion model to generate the corresponding image. Subsequently, a robust open-vocabulary detector [\citenum{b57}] is employed to assess whether the specified categories are accurately represented in the generated image. 1,000 text prompts are built by randomly sampling 80 categories of COCO instances [\citenum{b54}], which are used as inputs for the multi-category instance combinations. To avoid inference variance, each prompt is used to generate 10 rounds of images, resulting in a total of 10 × 1000 images. For each generated image, the open-vocabulary detector is used to determine how many different categories of objects. Based on detection results, MG2-5 metrics are defined through the overall success rates of generating 2-5 specified categories from 5 categories. \textbf{3) Fr$\acute{\textrm{e}}$chet Inception Distance (FID)} [\citenum{b59}] is used to assess the quality of the generated images from two datasets: i) 25014 image-caption pairs sampled from the COCO instance validation set; ii) 5000 image-caption pairs sampled from the Flickr30K instance validation set [\citenum{b60}]. \textbf{4) CLIP Score} [\citenum{b62}] is utilized to evaluate realism, which reflects the degree of match between generated images and text prompts. The evaluation datasets is the same as FID. \textbf{5) Efficiency} is used to compare the inference-time (i.e., the time required to generate an image using the trained model) of our ELDiff with other baselines. \textbf{5) T2I-CompBench} [\citenum{b63}] comprises 6,000 compositional text prompts evaluating 4 categories (attribute binding, object relationships, Numeracy, and complex compositions) and 7 sub-categories
(color binding, shape binding, texture binding, spatial relationships, non-spatial relationships, numeracy and complex compositions).

\section{Details of Finetuning and Inferencing}
\label{appD}
In \textbf{Algorithm 1}, We provide a detailed finetuning process for ELDiff, which addresses the segmentation map bias and semantic overlap conflict during the consistency constraint between the text prompts and the image contents, by pixel-level uncertainty estimation and token-level conflict optimization. The pseudocode for the denoising process of ELDiff is as follows.

\begin{algorithm}[H]
\caption{Finetuning for ELDiff}
\label{alg1}
\begin{algorithmic}[1]
\Require A text prompt $y$, a clear image $x_0$, and a pretrained stable diffusion model $\epsilon_{\theta}$.
\State $t\sim\operatorname{Uniform}(\{1,\ldots,T\})$
\State $\epsilon\sim\mathcal{N}(\mathbf{0},\mathbf{I})$
\State $c=f_\mathrm{{CLIP}}(y)$
\State $z_t=\sqrt{\bar{\alpha}_t}f_{VAE}(x_0)+\sqrt{1-\bar{\alpha}_t}\epsilon$
\For{$\mathrm{step} = 1, \ldots, S$}
    \State $\mathcal{L}_{\mathrm{LDM}}=\mathbb{E}_{z_0,t,c,\epsilon\sim\mathcal{N}(0,I)}\left[\left\|\epsilon-\epsilon_\theta\left(z_t,t,c\right)\right\|^2\right]$
    \For{$\mathrm{layer} = \mathrm{mid8}, \mathrm{up16}, \mathrm{up32}, \mathrm{up64}$}
        \State get the noun token's cross-attention maps $[A_{(1)}, \ldots, A_{(N)}]$ from each layer
        \State get the noun token's segmentation maps $[M_{(1)}, \ldots, M_{(N)}]$ from SAM
            \For{$(A_{(n)},M_{(n)}) = (A_{(1)},M_{(1)}), \ldots, (A_{(N)},M_{(N)})$}
            \State compute $\mathcal{L}_{\mathrm{PixEvi}}^{(n)}(A_{(n)}, M_{(n)})$ from Eq. 7
            \State compute $\mathcal{L}_{\mathrm{TokCon}}^{(n)}(A_{(n)}, A_{(n+1)})$ from Eq. 12
        \EndFor
    \EndFor
\EndFor
\end{algorithmic}
\end{algorithm}

\noindent
In \textbf{Algorithm 2}, We provide a detailed denoising process for ELDiff, which can be applied directly to the existing training pipeline of T2I diffusion models.

\begin{algorithm}[H]
\caption{Inferencing for ELDiff}
\label{alg1}
\begin{algorithmic}[1]
\Statex \textbf{Input:} A text prompt $y$ and the trained ELDiff
\Statex \textbf{Output:} A clear latent $z_0$
\State $z_T\sim\mathcal{N}(\mathbf{0},\mathbf{I})$
\State $c=f_\mathrm{{CLIP}}(y)$
\For{$t = T, \ldots, 1$}
\State $z_{t-1} = \frac{1}{\sqrt{\alpha_t}} \left( z_t - \frac{1 - \alpha_t}{\sqrt{1 - \bar{\alpha}_t}} \epsilon_\theta (z_t, t, c) \right)$
\EndFor
\State \Return $z_0$
\end{algorithmic}
\end{algorithm}

\section{Hyperparameter Ablation Study}
\label{appE}

\begin{table}[H]
\centering
\caption{Hyperparameter sensitivity analysis of $\lambda_{1}$, $\lambda_{2}$, $\eta_{1}$, and $\eta_{2}$.}
\label{tab:hyperparameter}
\scriptsize
\setlength{\tabcolsep}{2.5pt}
\renewcommand{\arraystretch}{1.05}

\begin{minipage}[t]{0.48\textwidth}
\centering
\textbf{(a) Results with different values of $\lambda_{1}$.}

\resizebox{\linewidth}{!}{
\begin{tabular}{c|ccccc}
\hline
$\lambda_{1}$ & OA$\uparrow$ & FID(C)$\downarrow$ & FID(F)$\downarrow$ & CLIP(C)$\uparrow$ & CLIP(F)$\uparrow$ \\ 
\hline
5e-4 & 0.7109 & 20.11 & 55.26 & 0.3229 & 0.3330 \\
5e-5 & 0.7116 & 20.09 & 55.12 & 0.3241 & 0.3348 \\
5e-6 & 0.7110 & 20.13 & 55.24 & 0.3237 & 0.3336 \\ 
\hline
\end{tabular}}
\end{minipage}
\hfill
\begin{minipage}[t]{0.48\textwidth}
\centering
\textbf{(b) Results with different values of $\lambda_{2}$.}

\resizebox{\linewidth}{!}{
\begin{tabular}{c|ccccc}
\hline
$\lambda_{2}$ & OA$\uparrow$ & FID(C)$\downarrow$ & FID(F)$\downarrow$ & CLIP(C)$\uparrow$ & CLIP(F)$\uparrow$ \\ 
\hline
5e-6 & 0.7105 & 20.17 & 55.33 & 0.3230 & 0.3328 \\
5e-7 & 0.7116 & 20.09 & 55.12 & 0.3241 & 0.3348 \\
5e-8 & 0.7097 & 20.14 & 55.15 & 0.3234 & 0.3343 \\ 
\hline
\end{tabular}}
\end{minipage}

\medskip

\begin{minipage}[t]{0.48\textwidth}
\centering
\textbf{(c) Results with different values of $\eta_{1}$.}

\resizebox{\linewidth}{!}{
\begin{tabular}{c|ccccc}
\hline
$\eta_{1}$ & OA$\uparrow$ & FID(C)$\downarrow$ & FID(F)$\downarrow$ & CLIP(C)$\uparrow$ & CLIP(F)$\uparrow$ \\ 
\hline
1e-3 & 0.7109 & 20.31 & 55.54 & 0.3226 & 0.3324 \\
1e-4 & 0.7116 & 20.09 & 55.12 & 0.3241 & 0.3348 \\
1e-5 & 0.7102 & 20.28 & 55.32 & 0.3230 & 0.3315 \\ 
\hline
\end{tabular}}
\end{minipage}
\hfill
\begin{minipage}[t]{0.48\textwidth}
\centering
\textbf{(d) Results with different values of $\eta_{2}$.}

\resizebox{\linewidth}{!}{
\begin{tabular}{c|ccccc}
\hline
$\eta_{2}$ & OA$\uparrow$ & FID(C)$\downarrow$ & FID(F)$\downarrow$ & CLIP(C)$\uparrow$ & CLIP(F)$\uparrow$ \\ 
\hline
1e-3 & 0.7103 & 20.46 & 55.44 & 0.3213 & 0.3320 \\
1e-4 & 0.7116 & 20.09 & 55.12 & 0.3241 & 0.3348 \\
1e-5 & 0.7088 & 20.38 & 55.65 & 0.3219 & 0.3314 \\ 
\hline
\end{tabular}}
\end{minipage}

\end{table}

\section{Results on SD v2.1 and SD v3.5 Medium}
\label{appF}

\begin{figure}[!htbp]
\centering
\includegraphics[width=0.8\linewidth]{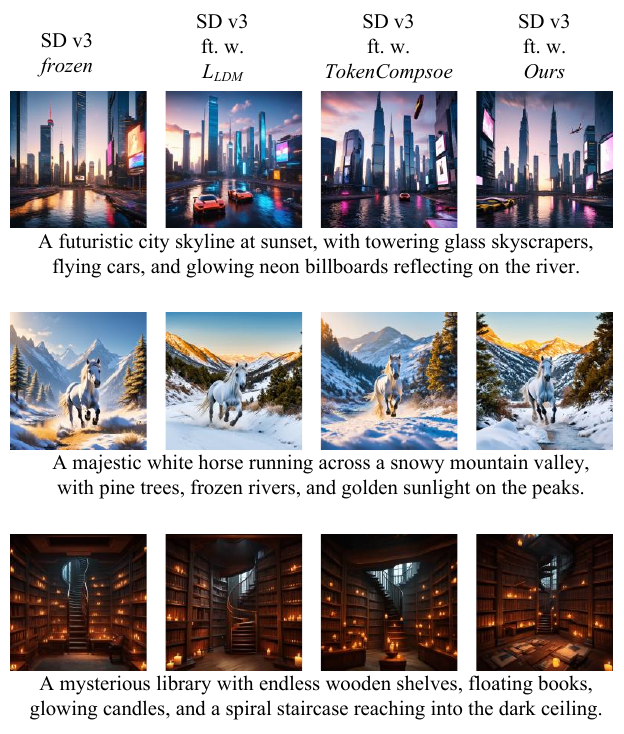}
\caption{Qualitative comparison of finetuning methods based on SD v3.5 Medium.}
\label{fig7}
\end{figure}

\label{appG}
\begin{figure}[!htbp]
\centering
\includegraphics[width=0.7\linewidth]{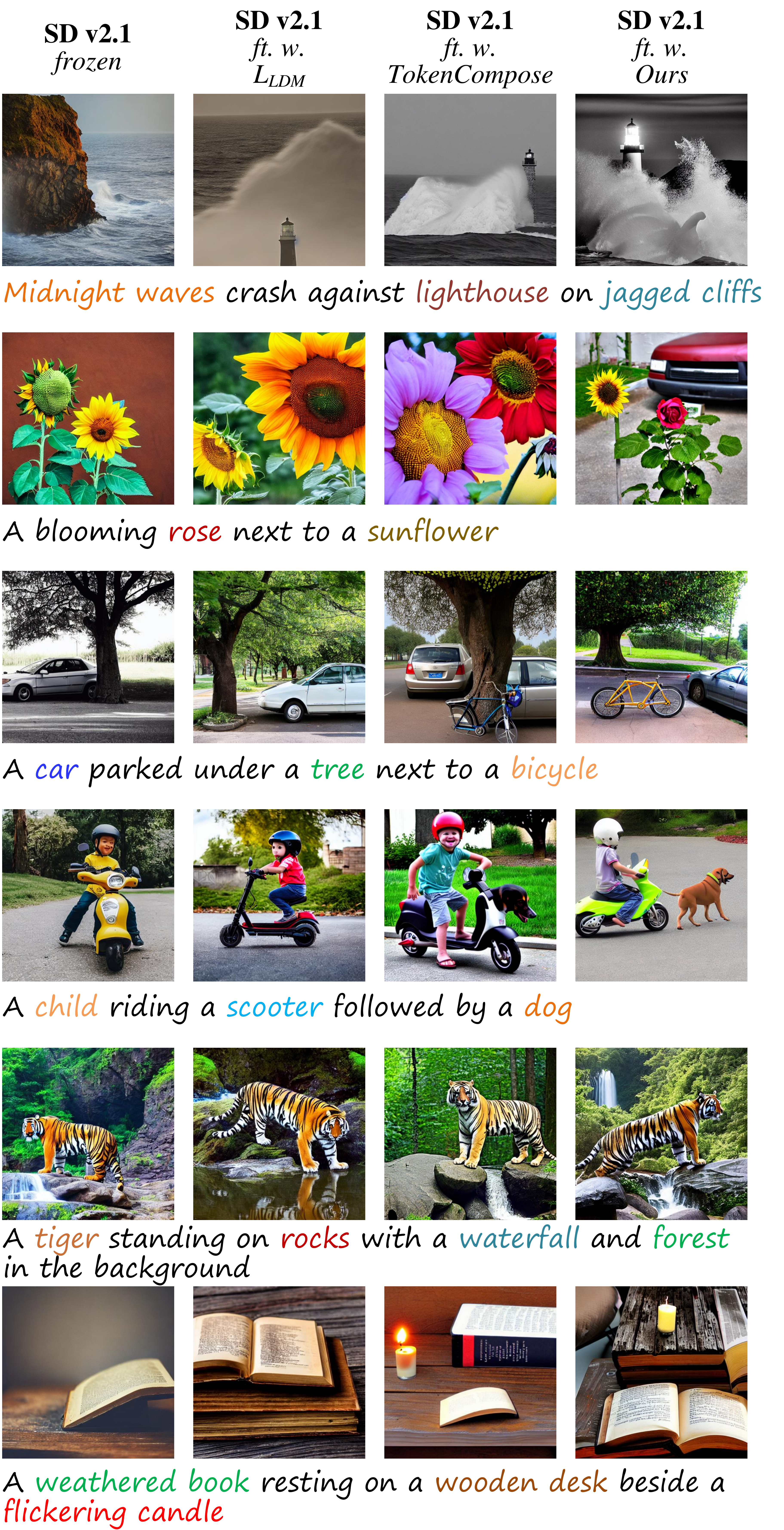}
\caption{Qualitative comparison of finetuning methods based on Stable Diffusion v2.1.}
\label{fig6}
\end{figure}

\end{document}